# Properties of simple sets in digital spaces. Contractions of simple sets preserving the homotopy type of a digital space.

Alexander V. Evako
Volokolamskoe Sh. 1, kv. 157, 125080, Moscow, Russia
Tel/Fax: +7 499 158 2939, e-mail: evakoa@mail.ru.

**Abstract**

A point of a digital space is called simple if it can be deleted from the space without altering topology. This paper introduces the notion simple set of points of a digital space. The definition is based on contractible spaces and contractible transformations. A set of points in a digital space is called simple if it can be contracted to a point without changing topology of the space. It is shown that contracting a simple set of points does not change the homotopy type of a digital space, and the number of points in a digital space without simple points can be reduces by contracting simple sets. Using the process of contracting, we can substantially compress a digital space while preserving the topology. The paper proposes a method for thinning a digital space which shows that this approach can contribute to computer science such as medical imaging, computer graphics and pattern analysis.

**Introduction**

Topological properties of two-and three-dimensional image arrays play an important role in image processing operations. A consistent theory for studying the topology of digital images in n dimensions can be used in a range of applications, including pattern analysis, medical imaging and computer graphics.
Usually, a digital object is equipped with a graph structure based on the local adjacency relations of digital points [1,6,10,20]. Transformations of digital objects preserve topological properties while changing the geometry of objects. One of the ways to do this in a small dimension is to use simple points: loosely speaking, a point of a digital object is called simple if it can be deleted from this object without altering topology. The detection of simple points is extremely important in image thinning, where a digital image of an object gets reduced to its skeleton with the same topological features. Local characterizations of simple points in three dimensions and efficient detection algorithms are particularly essential in such areas as medical image processing [2, 12,19], where the shape correctness is required on the one hand and the image acquisition process is sensitive to the errors produced by the image noise, geometric distortions in the images, subject motion, etc., on the other hand. The notion of a simple point was introduced by Rosenfeld [18]. Since then due to its importance, characterizations of simple points in two, three, and four dimensions and algorithms for their detection have been studied in the framework of digital topology by many researchers [3,5-8,11,13-14,16-18].
In papers [9,13-15], properties of contractible transformations of graphs based on deleting and gluing simple points and edges and preserving global topology of graphs as digital spaces were studied. In particular, it was shown that contractible transformations retain the Euler characteristic and homology groups of a graph. The outcome of the paper is based on computer experiments described in [14]. An obvious question arises from computer experiments: what is the way to reduce the number of points in a graph (digital object) which contains no simple point?
Section 2 of this paper describes computer experiments which lead to notion of a digital space as the digital model of a continuous object. A digital space is presented by a simple undirected graph. Sections 3 considers properties of contractible graphs, contractible transformations, simple points and edges of graphs and defines the homotopy equivalence of graphs. In section 4, we introduce the notion of a simple set of points and show that the contraction of a simple set converts a given graph to a homotopy equivalent graph with the same topological properties. We describe a method for transforming a digital space (graph) G to a homotopy equivalent skeleton which contains no simple point and no simple set.

## 2. Computer experiments

The following surprising fact was noticed in computer experiments described in [14]. Suppose that S is a surface in Euclidean space $E^n$. Divide $E^n$ into a set F of cubes with the edge length equal to L and vertex coordinates equal to nL. Call the cubical model of S the family M of cubes intersecting S, and the digital model of S the intersection graph G of M. Suppose that $S_1$ and $S_2$ are isomorphic surfaces, and $G_1$ and $G_2$ are their digital models.

It was revealed that there exists $L_0$ such that for any $L<L_0$, digital models $G_1$ and $G_2$ can be transformed from one to the other with some kind of transformations called contractible.

One can assume that the cubical and digital models contain topological and perhaps geometrical characteristics of the surface S. Otherwise, the digital model G is a discrete counterpart of a continuous space S [13-15].

To illustrate these experiments, consider examples depicted in fig. 1 and 2. In fig. 1, M, $M_1$ and $M_2$ are cubical models of circles, G(M), $G(M_1)$ and $G(M_2)$ are the intersection graphs of M, $M_1$ and $M_2$. G(M), $G(M_1)$ and $G(M_2)$ are homotopy equivalent to each other. $G(M_1)$ and $G(M_2)$ are digital 1-dimensional spheres.

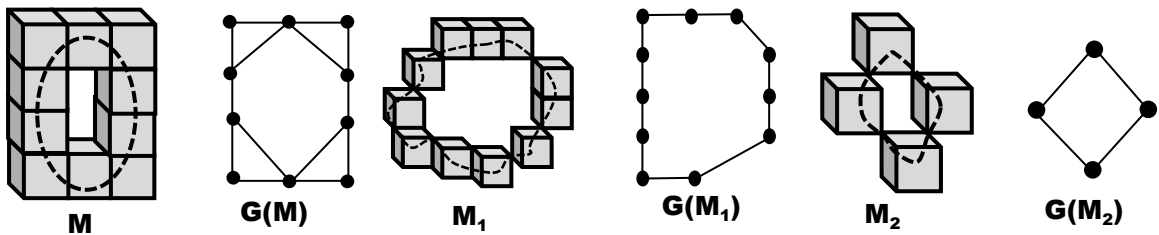

**Figure 1.** Cubical spaces M, $M_1$ and $M_2$ are cubical models of a circle. G(M), $G(M_1)$ and $G(M_2)$ are intersection graphs of M, $M_1$ and $M_2$. G(M), $G(M_1)$ and $G(M_2)$ are homotopy equivalent to each other. $G(M_1)$ and $G(M_2)$ are digital 1-dimensional spheres.

Cubical spaces M and $M_1$ in fig. 2 are cubical models of topological spheres in $E^3$, G(M) and $G(M_1)$ are the intersection graphs of M and $M_1$. It is easy to check that G(M) and $G(M_1)$ are homotopy equivalent to each other, i.e., G(M) can be converted to $G(M_1)$ by contractible transformations, and $G(M_1)$ is a minimal digital 2-sphere.

## 3. Contractible graphs, simple points and edges of graphs. Contractible transformations of graphs.

For the convenience of readers, we include some results related to contractible transformations of graphs. Proofs of basic properties of contractible graphs and contractible transformations can be found in papers [7-10, 13-15].

By a graph we mean a simple undirected graph G=(V,W), where V={$v_1,v_2,...v_n,...$} is a finite or countable set of points, and W = {$(v_pv_q),....$}⊆V×V is a set of edges. Such notions as the connectedness, the adjacency, the dimensionality and the distance on a graph G are completely defined by sets V and W.

Since in this paper we use only subgraphs induced by a set of points, we use the word subgraph for an

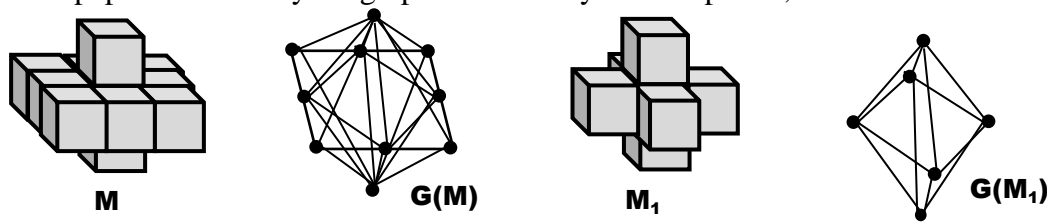

**Figure 2.** Cubical spaces M and $M_1$ are cubical models of a topological sphere. circle. G(M) and $G(M_1)$ are intersection graphs of M and $M_1$. G(M) and $G(M_1)$ are homotopy equivalent to each other. $G(M_1)$ is a minimal digital 2-sphere.

induced subgraph. We write H⊆G. Let G be a graph and H⊆G. G-H will denote a subgraph of G obtained from G by deleting all points belonging to H. For two graphs G=(X,U) and H=(Y,W) with disjoint point sets

X and Y, their join G⊕H is the graph that contains G, H and edges joining every point in G with every point in H. The subgraph O(v)⊆G containing all points adjacent to v (without v) is called the rim or the neighborhood of point v in G, the subgraph U(v)=v⊕O(v) is called the ball of v. Graphs can be transformed from one into another in a variety of ways. Contractible transformations of graphs seem to play the same role in this approach as a homotopy in algebraic topology [13-15].

**Definition 3.1.**
  (a) The trivial graph K(1) is contractible.
  (b) If H is a contractible subgraph of a contractible graph G then the graph G∪{v} obtained by gluing a point v to G in such a way that the rim O(v)=H is a contractible graph.
  (c) The family T={K(1), $G_1$, $G_2$, …$G_k$,…} of contractible graphs is determined by inductive application of operations (a)-(b).

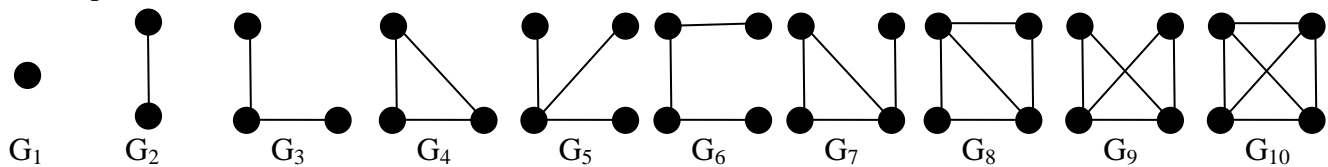

$G_1$  $G_2$  $G_3$  $G_4$  $G_5$  $G_6$  $G_7$  $G_8$  $G_9$  $G_{10}$

**Figure 3. Contractible graphs with the number of points n<5.**

**Definition 3.2.**
Let G be a graph, x be a point of G and (uv) be an edge of G.
- We say that x is a simple point if the rim O(x) of x a contractible graph.
- We say that (uv) is a simple edge if the joint rim O(uv)=O(u)∩O(v) is a contractible graph.

Figure 3 shows contractible graphs with the number of points k<5. $G_2$ is obtained from $G_1$ by gluing a simple point. $G_3$ and $G_4$ are obtained from $G_2$ by gluing a simple point. $G_5$ and $G_6$ are obtained from $G_3$ by gluing a simple point. $G_7$, $G_8$, $G_9$ and $G_{10}$ are obtained from $G_4$ by gluing a simple point. In graphs $G_4$, $G_9$ and $G_{10}$, any edge is simple.

Consider properties of contractible graphs. The following proposition is an evident consequence of definition 3.1.

**Proposition 3.1** ([7, 13-15]).
Let G∈T, H∈T and H⊆G.
- If H≠G. Then G-H contains a simple point.
- If H contains all simple points of G then H=G.

According to this proposition, if H contains all simple points of G then H=G. In fig. 4, a contractible graph G containes simple points {a,e,g}.

**Proposition 3.2.**
Let G be a graph and v be a point (v∉G). Then the cone v⊕G is a contractible graph.
**Proof.**
The proof is by induction on the number |G| of points of G. For |G|=1,2, 3,… the proposition is plainly true. Assume that the proposition is valid whenever |G|<k. Let |G|=k. Pick a point x∈G. The rim O(x) of x in v⊕G is v⊕$O_G$(x), where $O_G$(x) is the rim of x in G by construction of v⊕G. Since |$O_G$(x)|<k then v⊕$O_G$(x) is a contractible graph by the induction hypotheses. Therefore, x is a simple point of v⊕G. The graph H=(v⊕G)-x=(v⊕(G-x)) is contractible by the induction hypotheses. Therefore, v⊕G=H∪{v} is a contractible graph according to definition 3.1 . □

A contractible graph v⊕H is shown in fig. 4.

**Corollary 3.1.**

Let v be a point of a graph G. The ball U(v)={v}⊕O(v) of v is a contractible graph.

**Proposition 3.3.**
Let G and H be graphs. If G is a contractible graph then H⊕G is a contractible graph.
**Proof.**
The proof is by induction on the number |G| of points of G. For |G|=1,2, the proposition is plainly true. Assume that the proposition is valid whenever |G|<k. Let |G|=k. Pick a simple point x∈G. This means that the rim $O_G(x)$ of x in G is a contractible graph. The rim O(x) of x in H⊕G is H⊕$O_G(x)$. Since |$O_G(x)$|<k then H⊕$O_G(x)$ is a contractible graph by the induction hypotheses. For the same reason as above, the graph

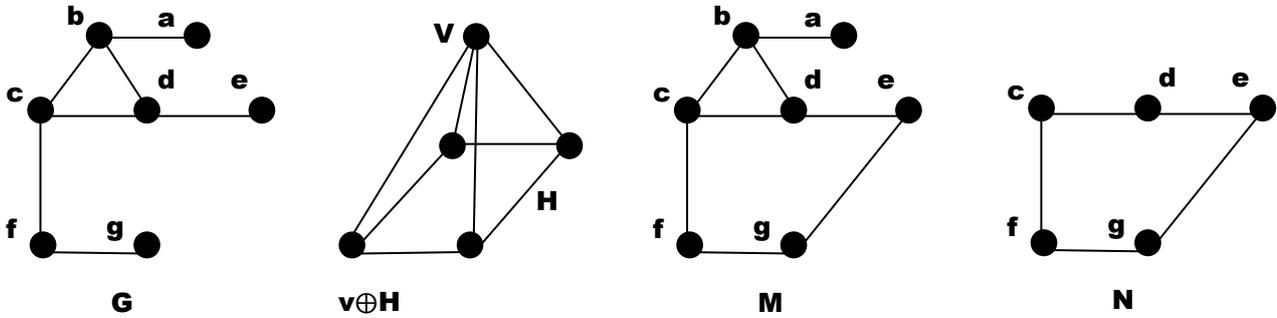

**Figure 4.** G is a contractible graph, {a,e,g} are simple points in G. v⊕H is a contractible graph. A graph M can be converted to N by sequential deleting simple points {a,b}.

P=(H⊕G)-x=(H⊕(G-x)) is contractible. Therefore, H⊕G=P∪{x} is a contractible graph. □

**Proposition 3.4.**
(a) Let G∈T, and |G|>1. Then G contains at least two simple points.
(b) Let G∈T, and a point v belong to G. Then G can be converted to v by sequential deleting simple points.

**Proof.**
(a) To prove (a), pick a point v belonging to G. Since {v} is a contractible graph then there is a simple point x belonging to G-v according to proposition 3.1. The graph {x} is a contractible subgraph of G. Therefore, there is a simple point y belonging to G-{x}. Thus, points x and y are both simple points of G. The proof is complete.
(b) Assertion (b) follows from proposition 3.1. □

Contractile graphs $G_2$-$G_7$ depicted in fig. 3 contain at least two simple points. Graph G shown in fig. 4 can be converted to a point d by sequential deleting simple points {a,b,e,g,f,c}.

**Proposition 3.5.**
Let G be a contractible graph and (uv) be a simple edge of G. Then the graph G-(uv) resulting from deleting edge (uv) from G is contractible.
**Proof.**
The proof is by induction on the number |G| of points of G. For |G|<5, the proposition is verified directly (see fig. 3). Assume that the proposition is valid whenever |G|<k. Let |G|=k. Consider the subgraph H=O(uv)∪{u,v} of G. H is a contractible graph according to proposition 4. Therefore, there is a simple point x of G belonging to G-H, and G-x is a contractible graph, |G-x|<k. Hence, P=(G-x)-(uv) is a contractible graph by the induction hypothesis. This means that P∪{x}=G-(uv) is a contractible graph. □

A contractible graph G in fig. 4 contains simple edges (bc), (bd) and (cd).

**Proposition 3.6.**
Let G be a contractible graph and v be a simple point of G. If |O(v)|>1, then G contains a simple edge.
**Proof.**

Suppose that v is a simple point, i.e., H=O(v) is a contractible graph and |O(v)|>1. Then H contains a simple point u. This means that the rim $O_H(u)$ of u in H is a contractible graph. By construction, $O_H(u)=O(u)\cap O(v)$. Hence, (uv) is a simple edge. □

**Definition 3.3.**
Deletions and attachments of simple points and edges are called contractible transformations. Graphs G and H are called homology equivalent if one of them can be converted to the other one by a sequence of contractible transformations.

Fig. 4 shows homotopy equivalent graphs M and N. M can be converted to N by sequential deleting simple points {a, b}. According to [7,10], N is a digital 1-sphere. It follows from the previous results that if graphs G and H are homotopy equivalent, and G is contractible, then so does H. It was shown in [14] that contractible transformations retain the Euler characteristic and the homology groups of a graph.

# 4. Contractions of simple sets of points

Good pairs of adjacency relations in grid-cell spaces were studied in [2]. Paper [17] has studied properties of non-trivial simple sets in the framework of cubical 3-D complexes. It is shown that the homotopy type of the object is not changed when the set is removed. Notice that in this paper we do not use simplicial or cubical complexes, we use an approach in which a digital object is represented by a graph. In paper [9], a simple pair of points in a graph was defined, and it was shown that contracting a simple pair preserves the homotopy type of a graph. In this section, we introduce extension of the notion of a simple point and a simple pair and a simple set of points. This allows reducing the number of points of a digital space which does not contain a simple point. In graph theory, the contraction of points x and y in a graph G is the replacement of x and y with a point z such that z is adjacent to the points to which points x and y

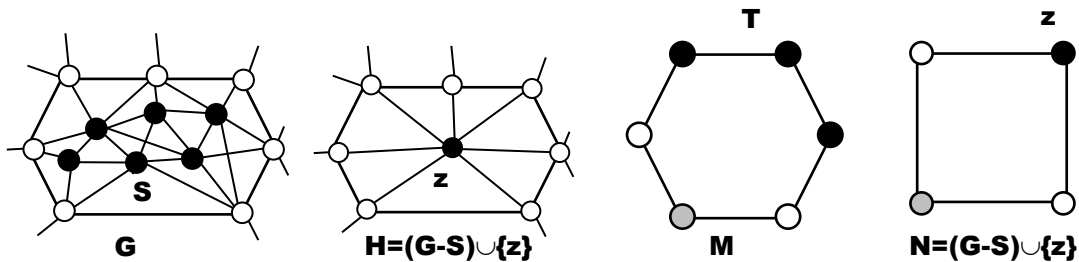

**Figure 5.** G is a graph, S is a simple set (black points), U(S) is a contractible graph (black and white points),. H=(G-S)∪{z} is obtained by the contraction of S. M is a digital 1-sphere, T is a simple set (black points), U(T) is a contractible graph (black and white points). A minimal digital 1-sphere N=(M-T)∪{z} is obtained by the contraction of T.

were adjacent.

**Definition 4.1.**
Let G be a graph and the set $S=\{v_1,v_2,…v_n\}$ be a contractible subgraph of G. S is said to be simple if the union $U(S)=U(v_1)\cup…U(v_n)$ is a contractible subgraph of G.

If $S=\{v_1,v_2\}$ then S is called a simple pair of points [9]. Notice that if G is a contractible graph with the set of points $V=\{v_1,v_2,…v_n\}$ then V is a simple set.

**Definition 4.2.**
Let G be a graph and $S=\{v_1,v_2,…v_n\}$ be a simple set of points in G. The contraction of set S to a point z in G is the replacement of points $\{v_1,v_2,…v_n\}$ with a point z such that z is adjacent to the all points to which points $\{v_1,v_2,…v_n\}$ were adjacent. By H=(G-S)∪{z} we denote the graph obtained by contracting S to z.

Fig. 5 illustrates definitions 4.1 and 4.2. In graph G, S is a simple set (black points), i.e., a contractible graph, U(S) is a contractible graph (black and white points), and H is the graph obtained by contracting S to z.

**Proposition 4.1.**
Let G be a contractible graph and S={$v_1,v_2,…v_n$} be a simple set of points of G. The graph H=(G-S)$\cup${z} obtained by contracting S to z is contractible.
**Proof.**
We have to show that G is converted to P by a sequence of contractible transformations.
The subgraph H=U($v_1$)$\cup$…U($v_n$) is a contractible graph. Glue a point z to G in such a way that O(z)=H. The obtained graph B=G$\cup${z} is homotopy equivalent to G, i.e., contractible. By construction of B, any point $v_i$ belonging to S is simple in B because the rim $O_B(v_i)$ of $v_i$ in B is the cone z$\oplus O_G(v_i)$, i.e., $v_i$ can be deleted from B. The obtained graph B-S=(G-S)$\cup${z} is homotopy equivalent to B, and therefore, to G. Thus, B-S is a contractible graph. □

**Proposition 4.2.**
Let G be a graph and S={$v_1,v_2,…v_n$} be a simple set of points in G. The graph H=(G-S)$\cup${z} obtained by contracting S to z is homotopy equivalent to G.
**Proof.**
Glue a point z to G in such a way that the rim O(z)=U(S). Since z is a simple point then the graph B=G$\cup${z} is homotopy equivalent to G. The rim $O_B(v_i)$ of $v_i$ in B is the cone z$\oplus O_G(v_i)$. Therefore, $v_i$ is a simple point of B and can be deleted from B. The obtained graph B-{$v_i$} is homotopy equivalent to B. For the same reason, graph B-S=(G-S)$\cup${z} is homotopy equivalent to B, and therefore, to G. □

Fig. 5 presents graph M which is a digital 1-sphere with six points. M has no simple point. T is a simple set of black points, U(T) is a contractible graph (black and white points) and N=(M$\cup$z)-T is a digital 1-sphere with four points. N is obtained by contraction of S. M and N are homotopy equivalent according to proposition 4.2.

Now we can describe a method for transforming a digital space (graph) G to a homotopy equivalent skeleton of G.
- Sequentially delete simple points from G. We obtain a compressed digital space $G_1$ with no simple point.
- Sequentially contract simple sets in $G_1$. Finally, we obtain a compressed digital space $G_2$ with no simple point and simple set of point.

It is clear, that realizing this method is a computational problem, which is understood to be a task that is in principle capable to being solved by a computer.

**Conclusion**
- A digital space (graph) can be converted to a skeleton (a graph with no a simple point and a simple set) by sequential deleting simple points and sequential contracting simple sets of points.
- Deleting a simple point and contracting a simple set of points preserves the homotopy type of a digital space.